# Real-time Prediction of Intermediate-Horizon Automotive Collision Risk

Blake Wulfe
Stanford University
Stanford, CA
wulfebw@stanford.edu

Sunil Chintakindi
The Allstate Corporation
Northbrook, IL
sunil.chintakindi@allstate.com

Sou-Cheng T. Choi
The Allstate Corporation
Northbrook, IL
soucheng.choi@allstate.com

Rory Hartong-Redden
The Allstate Corporation
Northbrook, IL
rory.hartong-redden@allstate.com

Anuradha Kodali
The Allstate Corporation
Northbrook, IL
akoda@allstate.com

Mykel J. Kochenderfer
Stanford University
Stanford, CA
mykel@stanford.edu

## ABSTRACT
Advanced collision avoidance and driver hand-off systems can benefit from the ability to accurately predict, in real time, the probability a vehicle will be involved in a collision within an intermediate horizon of 10 to 20 seconds. The rarity of collisions in real-world data poses a significant challenge to developing this capability because, as we demonstrate empirically, intermediate-horizon risk prediction depends heavily on high-dimensional driver behavioral features. As a result, a large amount of data is required to fit an effective predictive model. In this paper, we assess whether simulated data can help alleviate this issue. Focusing on highway driving, we present a three-step approach for generating data and fitting a predictive model capable of real-time prediction. First, high-risk automotive scenes are generated using importance sampling on a learned Bayesian network scene model. Second, collision risk is estimated through Monte Carlo simulation. Third, a neural network domain adaptation model is trained on real and simulated data to address discrepancies between the two domains. Experiments indicate that simulated data can mitigate issues resulting from collision rarity, thereby improving risk prediction in real-world data.

## KEYWORDS
Automotive Risk Prediction, Policy Evaluation, Monte Carlo Simulation, Bayesian Networks, Importance Sampling, Domain Adaptation

## 1 INTRODUCTION
The ability to accurately predict intermediate-horizon automotive risk from scene information is valuable for safety applications. For example, this capability can be used to inform the control-hand off problem in level-3 autonomous vehicles, or to allow for preemptive rerouting of autonomous vehicles away from high-risk situations.

There are three major challenges when predicting automotive risk from scene information. The first challenge is partial observability, which can arise due to occluded vehicles and sensor uncertainty along with unobserved driver information, such as degree of aggressiveness or maneuver intention. We assume full observability in this work for simplicity. The second challenge is that, as the prediction horizon increases, the set of possible vehicle trajectories grows exponentially. As a result, measures of risk associated with real-world data are generally imprecise, and estimating risk in simulation is computationally expensive. We restrict ourselves to an intermediate horizon in this paper.

The third challenge in intermediate-horizon risk prediction, and the focus of this paper, is a lack of data for fitting predictive models. This lack of data results from collision rarity and the high-dimensional nature of the problem. We demonstrate empirically that driver behavioral features have an increasingly significant impact as the prediction horizon increases. Learning a predictive model of risk that depends upon relatively high-dimensional behavioral features for many nearby vehicles requires a large amount of data with coverage over the state space, which in high-risk cases is unavailable. With collisions occurring approximately twice per million vehicle miles traveled in the U.S. [53], an economical and safe method for addressing collision rarity when fitting predictive models is desired.

A variety of methods have been considered for addressing the challenge of collision rarity in learning a predictive model. A common approach is to focus on predicting risk surrogates, such as hard braking or low time-to-collision events [10, 15]. While prediction of these quantities largely avoids issues resulting from rarity, it relies on the assumptions that these events correlate well with collisions and that they capture many collision modalities [18]. A second approach to addressing collision rarity is to augment existing data, for example through random noise or transforms [28], or more sophisticated oversampling [8]. These approaches typically assume smooth relationships between covariate and response variables, and it is not clear whether this holds in a risk prediction setting.

This paper aims to determine whether simulated data can help mitigate issues resulting from collision rarity. This approach leverages prior knowledge about the automotive domain, for example that vehicles behave according to physical laws or follow certain driver models, to generate sufficiently realistic data to improve prediction performance through transfer learning [38]. A simulation-based approach to addressing collision rarity faces two primary challenges: (i) efficient generation of high-risk simulated data, and (ii) reducing or compensating for inevitable differences between simulated and real-world data, thereby enabling effective transfer.

This paper presents a method that addresses these two challenges. The first challenge is addressed using importance sampling, which allows us to focus computational effort on higher-risk scenes. Determining proposal distributions from which to sample events can be challenging, so we propose an approach for automatically learning high-risk automotive scene distributions using the cross entropy method [11, 44]. We address the second challenge of effective transfer learning through a two-pronged approach. We first allow for simulated data to closely resemble real data when possible. We accomplish this by learning scene models from data, and by employing a risk estimation framework that imposes few assumptions on the dynamics or driver behavior models. Second, since simulated data will necessarily differ from real data, we use recent adversarial domain adaptation methods. These approaches have been shown to be effective in unsupervised settings [14], and we demonstrate their effectiveness in a practical, fully-supervised setting.

We validate our approach to intermediate-horizon risk prediction in two experiments. First, in a fully simulated setting, we demonstrate the effectiveness of the system in mitigating issues resulting from collision rarity. Second, we demonstrate that simulated data can be effectively transferred to improving a predictive model applied to real-world data.

## 2 RELATED WORK

Automotive risk models often address problems arising from collision rarity by predicting collision surrogates such as initiating conditions and evasive maneuvers [10, 15, 49]. These surrogate events are assumed to correlate with collisions [17, 20]. Lord et al. review approaches to statistical analysis of crash frequency data [33], which use *static* scene features. A variety of approaches have been considered in this context, including negative-binomial regression [34], support vector machines [32], and neural networks [7].

In contrast, real-time systems use *dynamic* features of a scene to estimate risk [31]. These approaches leverage some form of motion prediction, which range in complexity from physics-based models [3, 21] to those accounting for driver maneuver intention [30, 46]. Monte Carlo simulation methods permit complex dynamics and driver models by not making assumptions about these factors, and have been widely considered in automotive risk estimation [2, 6, 12].

Our framework uses a Monte Carlo approach, though it differs from existing methods in two important ways. First, previous research has emphasized short-term prediction, primarily with the goal of informing collision avoidance systems [31]. Second, whereas existing approaches propose to estimate risk in real-time by executing simulations on-vehicle, our approach amortizes the cost of running this optimization by learning a predictive model that generalizes across scenes.

Driver behavior models determine the actions taken by vehicles in automotive simulations. Heuristic driver models, such as the intelligent driver model (IDM) [51] and MOBIL [23], can be used to generate collision-free trajectories. A variety of collision-inclusive, heuristic models exist, for example, the Errorable Driver Model [57] and Less-Than-Perfect driver [56]. While these models exhibit collisions, truly human-like behavior and failure modes are better captured by fully parametric models that are learned from data

[29, 36]. We focus on the former heuristic models in this work, and leave more sophisticated models for future consideration.

Automotive scenes can be generated through heuristic means, or by learning a generative scene model from data. Wheeler et al. consider Bayesian networks [55] and factor graphs [?] for learning models directly from data. In this paper, we adapt the Bayesian network approach proposed by Wheeler et al. [55].

Due to collision rarity, many scene samples may be required to produce a collision, and many Monte Carlo simulations of those scenes may be needed to arrive at risk estimates with small relative error. Importance sampling provides a means of addressing these inefficiencies by oversampling dangerous scenes, and then weighing them based on their relative likelihood in an estimate or objective function [16]. Importance sampling has been applied for sampling lane-change scenarios, in particular with proposal distributions learned through the cross entropy method (CEM) [59]. CEM is a general optimization method common in rare event simulation [11, 44] that we also employ, but do so in learning full Bayesian network proposal distributions.

We use recent methods from the field of domain adaptation (DA) in order to improve transfer between simulated and real-world data. This problem has been considered in the context of reinforcement learning for robotic control [19, 45], as well as in supervised tasks such as pedestrian classification [54]. Unsupervised DA approaches typically seek to identify both domain invariant and discriminative features [14], or to infer shared and private latent spaces of the two domains [5], and have been applied, for example, to object classification [47].

## 3 PROBLEM STATEMENT

This section first introduces Markov decision processes and formulates risk estimation as policy evaluation. We then discuss three traits of the problem that will later motivate our approach.

### 3.1 Background

We formulate risk estimation within the Markov decision process (MDP) framework [4, 25, 41] due to its natural interpretation as policy evaluation. A finite-horizon MDP consists of a set of states $\mathcal{S}$, actions $\mathcal{A}$, a probability distribution $P(s' \mid s, a)$ over the next state $s'$ given the current state $s$ and action $a$, a reward function $R(s)$, which we assume is deterministic and a function of only the current state, a horizon $H < \infty$, and an initial distribution over states $\rho_1$.

A stochastic policy $\pi_\theta$ parameterized by $\theta \in \Theta$ defines a probability distribution over actions given the current state: $\pi_\theta(a \mid s)$. The return of a policy starting at time $t$ is the sum of rewards $r_t + r_{t+1} + \cdots + r_H$ it receives when interacting with an environment. The expected return of using policy $\pi_\theta$ starting from state $s_t$ is referred to as the value of the policy, which can be expressed using the Bellman equation:

$$V^{\pi_\theta}(s_t) = \mathbb{E}_{\pi_\theta}\left[\sum_{k=t}^{H} R(s_k) \,\middle|\, S_t = s_t\right]$$
$$= R(s_t) + \sum_a \pi_\theta(a \mid s_t) \sum_{s'} P(s' \mid s_t, a) V^{\pi_\theta}(s'). \quad (1)$$

Policy evaluation is the task of computing the value of a policy for all states. Policy evaluation can be performed through a dynamic programming procedure that iterates equation (1) as an update. This approach can be intractable in MDPs with large state or action spaces, in which case Monte Carlo policy evaluation might be employed instead, wherein sample returns are averaged to approximate the value of a policy.

## 3.2 Problem Formulation

We formulate automotive risk estimation as policy evaluation. A particular vehicle of concern, which we refer to as the ego vehicle, operates according to a policy $\pi_\theta$ parameterized by $\theta$. We refer to these parameters as the "behavioral features" of the vehicle because they are the parameters of the driver behavior models (e.g., IDM and MOBIL) that control vehicle action selection.

We assume full observability, and thus the state contains the physical attributes as well as the behavioral features of the ego vehicle and neighboring vehicles. Because we assume that behavioral traits are fully observed, this definition of the state corresponds most closely with the notion of a *scene* within the automotive safety literature [52], and we use "state" and "scene" interchangeably.

We are interested in automotive risk, which can be defined as the "likelihood and severity of the damage that a vehicle of interest may suffer in the future" [31]. We focus only on the probability of collisions, or occasionally on surrogate measures of risk, and therefore we would like to design a reward function such that the value of a policy corresponds to the probability of that particular policy suffering a collision. We accomplish this by defining an indicator function $C(s_t)$ equal to 1 if $s_t$ contains a collision involving the policy of interest, and 0 otherwise. We focus on collisions occurring after some initial timestep $h$, and define a collision indicator function that depends on this value:

$$Y_h(s_t) = \begin{cases} 1, & \text{if } C(s_t) = 1 \text{ and } t \geq h \\ 0, & \text{otherwise} \end{cases}. \quad (2)$$

States containing collisions of the ego vehicle are terminal states.

The value of a policy with respect to a state is the probability that policy enters into a collision starting in state $s_t$ and acting according to $\pi_\theta$ until the horizon $H$:

$$V^{\pi_\theta}(s_t) = \mathbb{E}_{\pi_\theta}\left[\sum_{k=t}^{H} R(s_k) \,\bigg|\, S_t = s_t\right] \quad (3)$$

$$= \sum_{k=t}^{H} \mathbb{E}_{\pi_\theta}\left[Y_h(s_k) \,\big|\, S_t = s_t\right] \quad (4)$$

$$= \sum_{k=t}^{H} P(Y_h(s_k) = 1 \mid S_t = s_t, \pi_\theta). \quad (5)$$

Above, we use the properties of linearity of expectation and the expectation of an indicator random variable. Defining states containing collisions as terminal states makes the events $Y_h(s_i) = 1$ and $Y_h(s_j) = 1$ mutually exclusive for all $i$ and $j$ such that $i \neq j$.

We define a trajectory as $\tau = (s_1, s_2, \ldots, s_H)$ in a space $\mathcal{T} = \mathcal{S} \times \cdots \times \mathcal{S}$, along with a function indicating that a trajectory contains a collision involving the ego vehicle:

$$Y_h(\tau) = \begin{cases} 1, & \text{if } C(s_t) = 1 \text{ and } t \geq h \text{ for any } s_t \in \tau \\ 0, & \text{otherwise} \end{cases}. \quad (6)$$

We can define the value as the probability of a collision in a trajectory from an initial state $s_t$. Hence (5) can be rewritten as

$$V^{\pi_\theta}(s_t) = P(Y_h(\tau) = 1 \mid S_t = s_t, \pi_\theta). \quad (7)$$

## 3.3 Traits of Intermediate-Horizon Automotive Risk Prediction

In this section, we discuss three traits of intermediate-horizon automotive risk prediction that will later inform our approach.

*3.3.1 Rarity of Collisions.* Validating autonomous systems frequently involves analyzing critical events that are rare. Formally, define the state-visitation distribution [43]

$$\rho(s_t) = \frac{1}{H}\sum_{k=1}^{H} P(S_k = s_t \mid \pi_\theta). \quad (8)$$

The expected value of a collision event, $\sum_{s_t} \rho(s_t)C(s_t)$, is assumed to be extremely small. This posses a challenge for learning to predict risk based on real-world data, and for Monte Carlo policy evaluation in simulation.

*3.3.2 State Space Size.* The state space in automotive risk prediction is high-dimensional and contains continuous variables. These factors motivate a method of generalizing risk estimates across scenes. Accomplishing this using function approximation implies learning a predictive model for risk conditioned on the scene.

*3.3.3 Influence of Behavioral Parameters.* In automotive risk prediction, behavioral parameters, $\theta$, of vehicles in the scene increase in importance with the prediction horizon. Figure 1 demonstrates this trend in an artificial [1] dataset. This fact, coupled with the relatively high-dimensional nature of behavioral features and the need to account for these features for all neighboring vehicles in the scene, makes intermediate-horizon automotive risk prediction a high-dimensional problem.

## 4 APPROACH

This section first introduces the simulator we use for data generation, which entails describing the components of the MDP. We then discuss our approach to policy evaluation (i.e., risk estimation). Finally, we describe how we fit a model to risk estimates in order to allow for real-time prediction.

### 4.1 Simulator

We define the simulator by specifying an initial scene distribution, $\rho_1$, and the transition function, $P(s' \mid s, a)$. This second component is composed of the physical dynamics model, and the driver behavior models that perform decision making for vehicles.

*4.1.1 Scene Generation.* Scene generation involves specifying the initial state distribution, $\rho_1$ of the MDP. As previously defined, the state includes the physical attributes and behavioral parameters, $\theta$, of vehicles in the scene. We assume a scene decomposes into $L$ individual vehicles as $S = \{S^{(1)}, S^{(2)}, \ldots, S^{(L)}\}$.

---
[1] We use artificial data due to the unavailability of collision-inclusive real-world data. Section 4.1 discusses the method used for generating this data in detail.

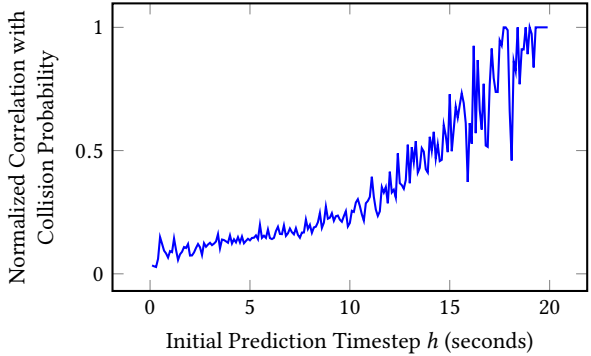

**Figure 1:** This plot shows, for a dataset of 70,000 vehicle trajectories each simulated 100 times, the maximum correlation across all behavioral features with rear-end collisions, normalized by the maximum correlation across all features as a function of the prediction horizon. The upward trend of this plot indicates that as $h$ increases, behavioral features become increasingly informative of collision risk.

**Table 1: Scene Vehicle Features**

| Feature Name | Symbol | Description |
| --- | --- | --- |
| fore distance | $s_f$ | distance to vehicle in front |
| fore velocity | $v_f$ | velocity of vehicle in front |
| relative velocity | $\Delta v$ | rear velocity minus fore velocity |
| length | $l$ | length of vehicle |
| width | $w$ | width of vehicle |
| attentiveness | $att$ | whether or not driver is attentive |
| aggressiveness | $agg$ | aggressiveness of driver |

A simple approach to scene generation is to select from a database of real scenes with inferred behavioral features. The problems with this approach are that (i) scene data must be available and (ii) only previously observed scenes can be sampled, meaning generalization to new roadways is not possible. Scenes may instead be initialized heuristically, or according to a constant configuration that is simulated for a burn-in period. While these approaches are simple, specifying rules for initial configurations can be challenging, and simulation-based approaches rely on driver behavior models to produce realistic distributions over scenes.

Alternatively, a generative model for scenes can be learned from data. This approach allows for generalization to new settings, and provides likelihood estimates of scenes. For these reasons, we employ a learned generative model in the form of a Bayesian network [27], which defines a probability distribution over a set of variables $S$ as a product of conditional probability distributions. The distribution of each variable is defined conditional on the values of its parent variables as defined by a directed acyclic graph. The distribution decomposes as $P(S) = \prod_{S^{(i)} \in S} P(S^{(i)} \mid \text{parents}(S^{(i)}))$.

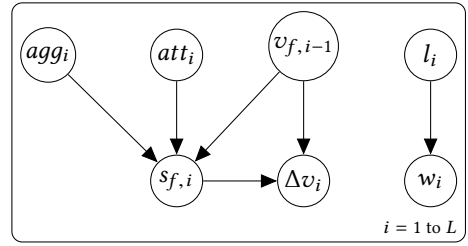

**Figure 2:** Bayesian network single-lane scene generation model.

Forward sampling allows for efficient generation of scenes from a Bayesian network provided that all observed nodes precede sampled nodes in a topological ordering [27]. We implement a single-lane, highway model similar to that of Wheeler et al. [55], with the primary difference being that we incorporate behavioral features in the model. Table 1 describes the variables of the network, and Figure 2 shows the plate model used for scene generation.

To sample a lane from the model, the variables of the first vehicle are sampled after marginalizing $v_f$. Subsequent vehicles are then sampled conditioning on the velocity of the prior vehicle. The scene distribution decomposes as

$$\rho_1(S) = P(S^{(1)}) \prod_{i=2}^{L} P(S^{(i)} \mid S^{(i-1)}). \qquad (9)$$

The scene model exclusively contains discrete variables. For sampling continuous values, the variables define bounds of a uniform distribution from which the value is sampled. This approach allows for the use of discrete variables, while still approximating arbitrary distributions as the discretization granularity increases, and has been used in the context of aircraft scene generation [26].

*4.1.2 Driver Behavior Models.* The previous section described how a scene is generated. We now discuss how this scene is simulated to produce vehicle trajectories. At each timestep, each vehicle samples a longitudinal and lateral acceleration determined by the driver models of that vehicle. For the longitudinal model, we employ a collision-inclusive variant of the IDM, based upon the Errorable Driver Model [57]. This model introduces collisions through a reaction time parameter that delays observations. The model also includes attentiveness parameters that determine the probability of a driver becoming distracted and not updating its action, as well as the probability of a driver becoming attentive from an inattentive state. We use MOBIL for lane changes, and sample longitudinal and lateral accelerations from Gaussian distributions. The means of these distributions are the acceleration values selected by the driver models, and the standard deviations are $0.5 \text{ m/s}^2$ and $0.1 \text{ m/s}^2$ for IDM and MOBIL, respectively.

For generating IDM and MOBIL parameters, we employ a correlated behavior model similar to that used in Sunberg et al. [48]. This approach samples, for each vehicle, an aggressiveness parameter from a uniform distribution over the unit interval. Aggressiveness then determines the mean of a truncated Gaussian distribution by interpolating between bounds for IDM and MOBIL as specified in Table 2. The standard deviation of this distribution is set to be 0.03 times the range of values. Given these actions, the simulator then propagates vehicles through space in discrete time increments of 0.1 seconds using a simple bicycle model [42].

Table 2: Driver model parameters and associated bounds

| IDM Parameter | Most Agg. | Least Agg. |
|---|---|---|
| maximum acceleration (m/s$^2$) | 6.0 | 2.0 |
| desired velocity (m/s) | 35 | 25 |
| min distance to fore vehicle (m) | 0 | 4 |
| safe time headway (s) | 0.2 | 1 |
| comfortable deceleration (m/s$^2$) | 5 | 2 |
| **MOBIL Parameter** | **Most Agg.** | **Least Agg.** |
| politeness | 0.1 | 0.5 |
| safe deceleration (m/s$^2$) | 2.0 | 2.0 |
| advantage threshold (m/s$^2$) | 0.01 | 0.7 |
| **Errorable Parameter** | **Value** | |
| reaction time (s) | 0.2 | |
| $p(\text{inattentive}_{t+1} \mid \text{attentive}_t)$ | 0.05 | |
| $p(\text{attentive}_{t+1} \mid \text{inattentive}_t)$ | 0.3 | |

## 4.2 Risk Estimation

This section discusses our approach to Monte Carlo policy evaluation, which we refer to as risk estimation in the automotive context.

*4.2.1 Monte Carlo Simulation.* The previous sections described the initial scene generation, driver behavior, and dynamics, which together specify the environment. Given this simulator, we can perform policy evaluation to estimate risk. We use Monte Carlo policy evaluation because it imposes minimal restrictions on the type of driver model used.

Our goal is to compute the value of a policy, which is the probability that that policy is involved in a collision conditional on a scene $s$ sampled from the scene model. Ideally, we would compute this probability exactly:

$$P(Y_h(\tau) = 1 \mid S = s, \pi_\theta) = \mathbb{E}_{\tau \sim P(T=\tau|S=s), \pi_\theta}[Y_h(\tau)]. \quad (10)$$

This value cannot be computed in closed form, so we instead approximate the value through sampling $n$ trajectories and averaging:

$$\mathbb{E}_{\tau \sim p(T=\tau|S=s), \pi_\theta}[Y_h(\tau)] \approx \frac{1}{n}\sum_{i=1}^{n} Y_h(\tau_i), \quad (11)$$

where $\tau_i \sim P(T = \tau \mid S = s_i, \pi_\theta)$.

*4.2.2 Importance Sampling.* Generating collision-inclusive data from the learned scene and driver models can require a large number of samples from (i) the scene generation model or (ii) the driver and dynamics models. We would like to reduce the quantity of samples needed so as to improve computational efficiency. Importance sampling samples collisions or other events with greater frequency relative to a baseline probability distribution, and then reweighs those samples according to their relative likelihood, thereby producing a lower variance, unbiased estimator [16]. This method can be applied in either of the sampling cases, but we focus on sampling initial scenes from the generative model.

The value in (10) conditions on a particular scene $s$. In contrast, our approach to importance sampling biases sampling towards scenes that are dangerous with respect to the ego vehicle. Formally, the challenge is that the unconditional probability of a collision

$$P(Y_h(\tau) = 1) = \mathbb{E}_{s \sim \rho_1(S=s), \tau \sim P(T=\tau|S=s), \pi_\theta}[Y_h(\tau)]$$

is small. We address this problem through importance sampling. Instead of sampling $s$ from $\rho_1(S)$, we sample from a proposal distribution $Q(S)$, which is biased towards dangerous scenes:

$$p(Y_h(\tau) = 1) = \mathbb{E}_{s \sim Q(S=s), \tau \sim P(T=\tau|S=s), \pi_\theta}\left[\frac{\rho_1(S=s)}{Q(S=s)} Y_h(\tau)\right].$$

This method has been employed in evaluating the safety of autonomous systems, for example, vehicles [59] and aircraft [24]. In these settings, importance sampling results in a lower variance estimator of the *unconditional* probability of collision or other event.

In our setting, we are *not* interested in computing the *unconditional* probability of a collision. Instead, we are interested in fitting a predictive model to the *conditional* probability of a collision with the goal of achieving high performance on certain evaluation metrics. Due to this altered goal, importance sampling in this context can be interpreted as a heuristic approach to active learning [9]. From this perspective, we make the assumption that dangerous scenes will improve performance by subjecting the model to a greater number of positive-risk events, thereby addressing the low sample size problem associated with those events.

*4.2.3 Cross Entropy Method.* A remaining task is determining the proposal distribution $Q(S)$. For this distribution, we use another Bayesian network, the conditional probability distributions of which have been altered to generate collisions more frequently. Because a scene filled with dangerous vehicles would result in extremely low likelihood values, we instead sample a single vehicle in each scene from $Q$, and the remaining vehicles as usual from $\rho_1$. Recall that the scene model decomposes across the $L$ vehicles as (9). The proposal distribution $Q(S)$ differs from $\rho_1(S)$ only in the $j$th vehicle sampled, and thus all vehicle probabilities cancel in the ratio of $\rho_1$ and $Q$ except for the $j$th vehicle:

$$\frac{\rho_1(S)}{Q(S)} = \frac{P(S^{(j)} \mid S^{(j-1)})}{Q(S^{(j)} \mid S^{(j-1)})}.$$

We refer to this likelihood ratio as $w$.

We could manually alter the Bayesian network parameters to increase the likelihood of collisions involving the ego vehicle, but we would like to automate this process. We accomplish this using the cross entropy method (CEM).

CEM [44] is an optimization algorithm originally designed for finding proposal distributions in the context of rare events, which operates by iteratively sampling from some distribution, and then updating the parameters of that distribution based on the fitness of the samples [11]. In the particular instantiation of CEM we employ, the proposal distribution is trained to produce collisions involving the ego vehicle within 10 to 20 seconds by changing the conditional probability distributions of the variables in the Bayesian network.

Figure 3 shows the mean values of three variables and collision probability during the optimization process. The changes to these variables, which may initially seem counterintuitive, result in a higher probability of collision. The reason for this is that only intermediate-horizon collisions (i.e., those occurring in the 10 to 20 second range) are counted. Thus, the method learns to avoid early collisions (i.e., those occurring before 10 seconds) by making the ego vehicle both attentive and slower than the vehicle in front. Because the ego vehicle initially travels at a lower velocity than the fore vehicle, it tends to speed up. If the ego vehicle then becomes inattentive, it will continue accelerating until colliding with the fore vehicle. Aggressiveness decreases because this reduces the initial acceleration (thereby preventing early collisions), as well as the magnitude of the comfortable deceleration rate (thereby limiting the ability of the ego vehicle to slow down).

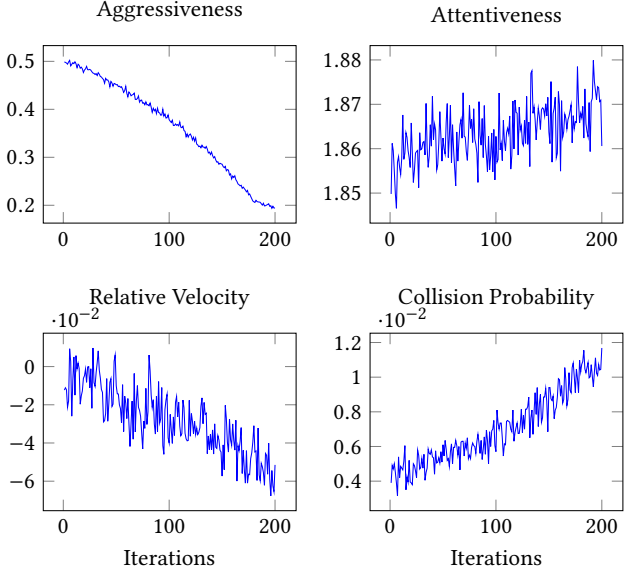

**Figure 3: Mean values for three of the Bayesian network variables and collision probability throughout optimization.**

A dataset of scenes is sampled from the learned distributions $\rho_1$ and $Q$, and each scene evaluated using Monte Carlo simulation. The resulting dataset $\{s^{(i)}, y^{(i)}, w^{(i)}\}_{i=1}^{m}$, containing $m$ scene-risk pairs, each with an associated likelihood ratio $w$, is then used to augment real data in learning a predictive model.

## 4.3 Risk Prediction

Risk estimation must be performed in real-time on-vehicle. A common approach to this task is to simulate future trajectories online to derive a risk estimate [2, 39, 58]. While this method may be feasible for short-term risk estimation, we believe that intermediate and longer-horizon risk will be impractical to estimate in this manner. There are two reasons why this would be the case. First, because driver behavior increases in significance with longer horizons, effective simulators will likely use learned, computationally expensive driver models [29]. Second, and more challenging, is the number of simulations that will need to be run due to the prediction horizon in order to arrive at estimates of risk with low relative error. We propose to instead fit a predictive model offline, the online computational complexity of which grows slowly as a function of simulator complexity and the number of simulations run. The following sections describe this prediction model, and our method for compensating for discrepancies between simulated and real-world data.

*4.3.1 Prediction Model.* The response variable $Y$ in collected datasets is the estimated proportion of successes $p$ (e.g., collisions) in a binomial experiment with $n$ samples. Because this value is estimated from a finite number of samples, it can only assume a finite number of values, and is therefore a discrete variable. This variable has two unusual properties. First, valid values are bounded between 0 and 1 inclusively. Second, in real-world data, predicting this variable reduces to binary classification.

Zhao et al. [60] compare linear and logistic regression in fitting proportion data, finding logistic regression to have better performance [60]. Three alternative approaches are (i) to take $\log(Y)$ as the response and fit a linear regression model, thereby avoiding issues due to bounded values; (ii) to perform beta regression of the values, which requires a transform of response values of 0 or 1 [13]; and (iii) to treat the response as counts and fit a negative binomial regression [40], which would complicate transfer to real-world data where the sample size $n$ differs from simulation. While these alternatives may have merit, we elect to use the cross entropy loss because of its simplicity and good performance in initial experiments.

In risk *estimation*, we considered the risk associated with the ego vehicle in a scene $s$. In risk *prediction*, we take the perspective of the ego vehicle directly to reflect the information available in a real-world situation. We refer to the features from this perspective with the random variable $X$, which we assume is still a Markovian representation of the environment.

We fit a predictive model $V(x; \phi)$ with parameters $\phi$ as the value function, minimizing the loss

$$\mathcal{L}_{pred}(X, Y)$$
$$= -\mathbb{E}_{(x,y)\sim(X,Y)}[y \log(V(x;\phi)) + (1-y)\log(1-V(x;\phi))]. \quad (12)$$

We empirically evaluate this loss on the collected dataset as

$$-\sum_{i=1}^{m} w^{(i)}[y^{(i)}\log(V(x^{(i)};\phi)) + (1-y^{(i)})\log(1-V(x^{(i)};\phi))]. \quad (13)$$

This model is fit to a dataset containing risk estimates for many different policy parameters, $\theta$. We include these parameters as features, effectively learning a value function over a distribution of policies. In all experiments, we use a neural network prediction model with sigmoid activation function. This choice of model is motivated by the high-dimensional feature space, and the ability to employ recent domain adaptation approaches as we discuss next.

*4.3.2 Domain Adaptation.* Even in high-fidelity simulations, generated data will necessarily differ from real-world data. Formally, the joint probability distribution over the simulated, ego-centric scene representations $X_s$ and risk estimates $Y_s$, will differ from the (target) real-world joint distribution: $P_s(X_s, Y_s) \neq P_t(X_t, Y_t)$. This "domain shift" results in degraded performance when transferring from the source to target domain [45].

Domain adaptation (DA) methods attempt to address this problem. Both supervised and unsupervised variants exist, and it is not immediately clear which approach is better suited for automotive risk prediction. The reason for this is that the target, real-world automotive domain can be viewed as "weakly supervised" in the sense that its risk estimates are highly imprecise reflections of the true underlying collision probability.

Table 3: Prediction Features

| Feature Name | Example Quantities |
| --- | --- |
| Ego physical | lane offset and heading, velocity, vehicle length and width, acceleration, turn rate, time-to-collision |
| Ego well-behaved | ego vehicle currently colliding, out of lane, or has a negative velocity |
| Ego behavioral | IDM and MOBIL parameters |
| Neighboring vehicle state | same physical and behavioral features as as ego vehicle and relative dist. to ego |

While supervised DA methods have been shown to be effective, for example, the "semantic alignment loss" of Motiian et al. [37], we focus on an unsupervised approach, domain adversarial neural networks (DANN) [14]. DANN attempts to learn features that are (i) invariant to the domain and (ii) useful for the task being performed. To accomplish this task, the activations of certain layers in the network, $M_s$ and $M_t$, are encouraged to match across domains. This requires a measure of similarity, and for that DANN employs a domain classification network, $D$, that attempts to distinguish between the learned features of the two domains. The risk prediction network incurs an additional adversarial loss

$$\mathcal{L}_{adv}(X_s, X_t, M_s, M_t) \qquad (14)$$
$$= -\mathbb{E}_{x_s \sim X_s} [\log D(M_s(x_s))] - \mathbb{E}_{x_t \sim X_t} [\log(1 - D(M_t(x_t)))],$$

with $D$ being trained to minimize the negative. The overall objective then becomes $\mathcal{L} = \mathcal{L}_{pred} + \lambda \mathcal{L}_{adv}$, where $\lambda$ controls the weight of the adversarial loss.

Minimizing this loss with respect to both $M_s$ and $M_t$ encourages the source and target distributions to match, but this might come at the cost of learning a good target feature distribution. In the supervised setting, it is possible to only optimize the adversarial loss with respect to the source feature distribution $M_s$, and we consider this approach as well in our experiments.

## 5 EXPERIMENTS AND RESULTS

Our goal is to assess whether simulated automotive data can improve intermediate-horizon risk prediction in real-world data. We perform two experiments. The first experiment employs the full approach as described, and seeks to determine whether importance sampled collision events can improve prediction in an artificial scenario. The second experiment considers the simulated to real transfer challenge using real-world vehicle trajectory data.

Throughout the experiments, we use a set of scene features, which we summarize in Table 3. Our primary focus in this research is on addressing the challenge of collision rarity, but not on partial observability. As a result, we assume access to features that are difficult to obtain in real scenarios, for example, information about vehicles far in front of the ego vehicle, and behavioral features that typically must be inferred [48].

### 5.1 Simulation Experiments

In this section, we assess whether importance sampling of collisions in artificial settings can improve prediction performance. Our focus on exclusively simulated data in this experiment is motivated by the unavailability of real-world data containing collisions.

We generate an artificial dataset to act as the "real" data containing few collisions. We generate this data heuristically, randomly initializing vehicles and their behavioral parameters on a single lane, circular track, which is simulated for 600 timesteps to arrive at a random configuration. This configuration is then simulated 100 times for 200 timesteps, and risk estimates for the period between 100 and 200 timesteps are computed. Each scene involves 70 vehicles, and we take each vehicle as a single sample. This produces a dataset in which samples are not independent since they may both be from the same scene. For the sake of efficiency, we ignore this lack of independence in the training data; however, the validation set contains samples generated from entirely different simulations from those of the training set.

We apply the proposed system by fitting a scene model, running the cross entropy method to derive a proposal distribution, generating a dataset, and learning predictive models. We compare four methods: (i) training only on the target domain, (ii) training on both domains without adaptation, (iii) training on both domains with adaptation, and (iv) training on both domains, applying adversarial loss only to the source features. Because the number of collision events in the target domain plays a critical role in determining the value of simulated data, we compare performance between the models for various numbers of target domain collision events. To better reflect real-world data, we convert the target *training* set into binary values by sampling collisions according to the observed probabilities. The validation set we leave as continuous values because this provides greater precision during evaluation.

A hyperparameter search was performed for each method over a predefined set of three network architectures, learning rates between $10^{-4}$ and $10^{-3}$, and dropout probabilities between 0.5 and 1. The three architectures consisted of encoding hidden layers containing (512, 256, 128, 64), (256, 128, 64), or (128, 64) units, and classifier hidden layers containing either no hidden units (i.e., direct classification of encoder output), (64) or (64, 64) hidden units. All networks used ReLU activations, and for each training run the best validation performance during the run was selected for use in the results. The domain adversarial loss was applied to the features output from the encoder. Thirty networks were trained for each method. Figure 4 (a) shows the average negative log likelihood evaluated on the validation set across these training runs.

These results support two main conclusions. First, all models that learn from simulated data exhibit improved performance on positive-risk instances, and source-only adaptation and no adaptation improve over target-only training overall. This result is likely due to the fact that the jointly-trained models observe a far greater number of positive-risk samples from the source dataset, thereby enabling them to learn a better predictive model. Second, when few positive-risk samples are provided in the training set, the adaptation models achieve the best performance. With more positive-risk samples the model without adaptation performs best across all validation samples, likely because this model is better able to decide what source data to leverage, rather than being explicitly constrained to learning a shared feature space.

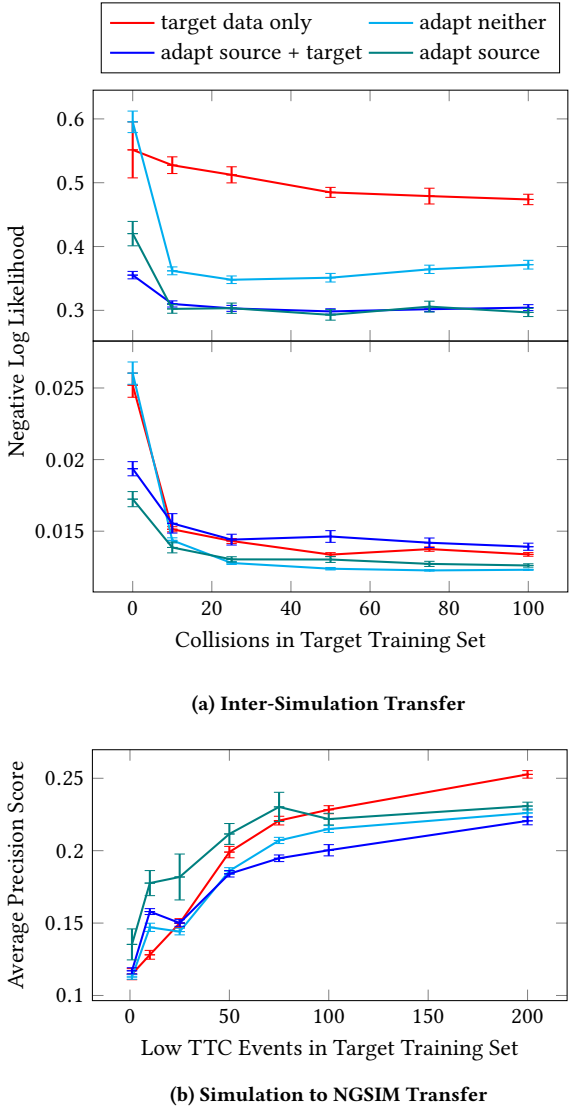

(a) Inter-Simulation Transfer

(b) Simulation to NGSIM Transfer

**Figure 4:** These plots show validation results with varying amounts of *positive*-risk samples in the target domain training set. Each point represents the mean performance across runs, with vertical bars indicating standard error. The top and bottom plots of (a) show performance evaluated on *positive*-risk and all target domain validation samples, respectively. Plot (b) shows results for transferring from simulation to NGSIM data.

### 5.2 Real-world Transfer Experiments

We next seek to determine whether simulated automotive data can be transferred to a prediction task involving real-world data. We consider the task of predicting low time-to-collision (TTC) events [15], which we define as a vehicle having a TTC less than 3.0 seconds, in the US Highway 101 portion of the Next-Generation Simulation (NGSIM) dataset [1]. NGSIM contains vehicle trajectories on a five-lane highway, collected over three 15-minute time periods. These trajectories contain no collisions, so we instead focus on a collision surrogate. Furthermore, low TTC events are not sufficiently rare to merit application of importance sampling, so we generate 50,000 source samples heuristically using the same method as used in the simulated experiments for mimicking real-world data. Because NGSIM contains binary labels, we now evaluate classification performance using the average precision score [61], which we choose in order to better account for class imbalance.

The NGSIM dataset does not contain explicit behavioral parameters. Since these parameters are central to the high-dimensionality of the risk prediction problem, we first extract IDM parameters using the local maximum likelihood method of Kesting et al. [22]. We smooth the U.S. Highway 101 subset of the NGSIM data according to the method of Theimann et al. [50], and then extract samples from it in 20 second increments. To minimize dependence between training and validation datasets, we train on the first and third time periods of the NGSIM dataset, and validate on the second.

We again compare performance of the four methods across varying amounts of positive-risk target data, and visualize results in Figure 4 (b). These results indicate that when training data is limited, simulated data can significantly improve prediction performance. As the number of positive-risk target samples increases, the target-only model performs best because enforcing domain-invariance begins to outweigh the benefits of additional positive-risk samples from simulated data. Approximately 5% of the samples have a positive response value. In this imbalanced setting, a random classifier achieves an average precision score of 0.05. All models therefore perform significantly above random, but nevertheless quite poorly due to the high-variance nature of the *classification* task.

## 6 CONCLUSION

This paper aimed to answer the question of whether simulated data could be used to improve automotive risk prediction. We concluded that simulated data can be helpful in cases where the amount of data available is insufficient to learn a good predictive model. We demonstrated that intermediate-horizon prediction suffers from this low sample size issue due to the significance of high-dimensional behavioral features in learning a predictive model. We then showed that domain adaptation methods could be used with simulated data to improve prediction in both artificial and real-world settings. The significance of this result is that systems that perform intermediate-horizon risk prediction may potentially be improved through the offline training of predictive models on simulated data.

Our approach can be improved through more sophisticated modeling, estimation, and prediction. Initial scene generation could employ the factor graph approach of Wheeler et al. [? ] or deep generative models. Driver models learned from data [29] are likely to provide substantial benefit in transfer performance by better capturing human failure modes. A multi-lane approach to importance sampling would allow for application of the system to real-life collision prediction transfer.

Prediction performance could be enhanced through the use of domain adaptation models that explicitly model shared and private latent feature spaces [5]. Finally, the local maximum likelihood estimation approach to inferring behavioral parameters of NGSIM vehicles is limited because the model poorly captures individual driving behavior, and it does not generalize across drivers. A method proposed by Morton et al. [35] would allow for arbitrarily complex behavior encodings, while also generalizing across samples, thereby alleviating issues resulting from the lack of individual-vehicle data.

# 7 ACKNOWLEDGMENT

The authors wish to thank Gina Madigan, Tim Wheeler, Zach Sunberg, and Katie Driggs-Campbell for helpful conversations and feedback. This research was funded by The Allstate Corporation.